# Evidential Reasoning with Conditional Belief Functions

Hong Xu*†                                    Philippe Smets*

IRIDIA* and Service d'automatique†
Université libre de Bruxelles
50, Ave. F. D. Roosevelt, CP 194/6, B-1050, Brussels, Belgium
Email: {hongxu, psmets}@ulb.ac.be

## Abstract

In the existing evidential networks with belief functions, the relations among the variables are always represented by joint belief functions on the product space of the involved variables. In this paper, we use conditional belief functions to represent such relations in the network and show some relations of these two kinds of representations. We also present a propagation algorithm for such networks. By analyzing the properties of some special evidential networks with conditional belief functions, we show that the reasoning process can be simplified in such kinds of networks.

## 1. INTRODUCTION

Network-based approaches have been widely used for knowledge representation and reasoning with uncertainties. Bayesian networks (Pearl 1988) and valuation network (Shenoy 1992) are two well-known frameworks for the graphical representations. Bayesian networks are implemented for the probabilistic inference, while valuation networks can represent several uncertainty formalisms in a unified framework. Graphically, a Bayesian network is a directed acyclic graph, a valuation network is a hypergragh. Nodes in the networks represent random variables where each variable is associated with a finite set of all its possible values called its frame. In a Bayesian network, arcs represent the relations among the variables in the form of conditional probabilities, in a valuation network, such relations are represented in the forms of joint valuations on the product space of the involved variables. For the case of belief functions, such valuations are the joint belief functions. Recently, Cano et al. (1993) have presented an axiomatic system for propagating uncertainty (including belief functions) in Pearl's Bayesian network, based on Shafer-Shenoy's axiomatic framework (Shafer and Shenoy 1988, Shenoy and Shafer 1990). But the belief functions for representing relations of the variables in their system are still represented on the product space. Smets (1993) has generalized the Bayes' Theorem for the case of belief functions and presented the Disjunctive Rules of Combination for two distinct pieces of evidence, which makes it possible for representing knowledge and reasoning in evidential network in the form of conditional belief functions. In this paper, we show that any joint belief function representing conditional relations can always be represented by a form of conditional belief functions. We then present a propagation scheme for more complicated cases of evidential networks proposed by Smets (1993). Specifically, we show that the reasoning process can be simplified in some special cases.

The rest of the paper is organized as follows: In section 2, we briefly review belief functions and their rules of combination, both conjunctive and disjunctive; In section 3, we show the relations between the joint belief functions and conditional belief functions which represent the same knowledge; In section 4, we first introduce the evidential network with conditional belief functions, next we present a propagation scheme for it, finally we analyze the properties of some special network and show how to simplify the computation in such networks; Finally in section 5, we give some conclusions.

## 2. BELIEF FUNCTIONS AND THEIR RULES OF COMBINATIONS

In this section, we briefly review the concept of belief functions (Shafer 1976, Smets 1988), and summarize the conditioning and combination rules for belief functions. More details can be found in (Smets 1990, 1993).

**Definition 1:** Let $\Omega$ be a finite non-empty set called *the frame of discernment* (*the frame* for short). The mapping bel: $2^\Omega \to [0, 1]$ is *an (unnormalized) belief function* iff there exists a *basic belief assignment (bba)* m: $2^\Omega \to [0, 1]$ such that:

$$\sum_{A \subseteq \Omega} m(A) = 1, \quad bel(A) = \sum_{B \subseteq A, B \neq \emptyset} m(B), \quad \text{and } bel(\emptyset) = 0.$$

*A vacuous belief function* is a belief function such that $m(\Omega) = 1$ and $m(A) = 0$ for all $A \neq \Omega$.

For a given belief function, we can define a *plausibility function* pl: $2^\Omega \rightarrow [0, 1]$ and a *commonality function* q:$2^\Omega \rightarrow [0, 1]$ as follows: for $A \subseteq \Omega$, $A \neq \emptyset$,

$$pl(A) = bel(\Omega) - bel(\overline{A}) \quad \text{and} \quad pl(\emptyset) = 0$$
$$q(A) = \sum\{m(B) \mid A \subseteq B \subseteq \Omega\}$$

where $\overline{A}$ is the complement of A relative to $\Omega$.

**Definition 2:** Let bel be our belief about the frame $\Omega$. Suppose we learn $\overline{A} \subseteq \Omega$ is false. The resulting *conditional belief function* bel$(.|A)$[1] (bel(B|A)) can be read as the belief of B given A) is obtained through *the unnormalized rule of conditioning*: for $B \subseteq \Omega$,

$$m(B|A) = \sum_{X \subseteq \overline{A}} m(B \cup X) \quad \text{if } B \subseteq A \subseteq \Omega$$
$$= 0 \quad \text{otherwise}$$

**Definition 3:** Consider two distinct pieces of evidence on $\Omega$ represented by belief functions bel$_1$ and bel$_2$. The belief function bel$_{12}$ that quantifies the combined impact of these two pieces of evidence is obtained through *the conjunctive rule of combination*. We use $\oslash$ to represent the conjunctive combination operator. $\forall A \subseteq \Omega$,

$$m_{12}(A) = \sum_{A = B \cap C} m_1(B) m_2(C). \tag{1}$$

It can also be represented as:

$$m_{12}(A) = \sum_{B \subseteq \Omega} m_1(A|B) m_2(B) \tag{2}$$

**Definition 4:** Consider two distinct pieces of evidence on $\Omega$ represented by belief functions bel$_1$ and bel$_2$. The belief function bel$_{12}$ induced by the disjunction of these two pieces of evidence is obtained through *the disjunctive rule of combination* (Dubois and Prade 1986). We use $\oslash$ to represent the disjunctive combination operator. $\forall A \subseteq \Omega$,

$$m_{12}(A) = \sum_{A = B \cup C} m_1(B) m_2(C) \tag{3}$$

Since m, bel, pl and q are in one-to-one correspondence with each other, the above rules can also be represented by using any of these functions. In this paper, we only give the formulas which will be used in the later computation.

Note that all the definitions above are for the non-normalized case. As for the case of normalized belief functions, which means $m(\emptyset)=0$, the normalization factor $K=1-m(\emptyset)$ should be considered in those rules, and the conditioning rule and the conjunctive combination rule turn out to be Dempster's rule of conditioning and of combination, (unnormalized) bel(A|B) turns out to be (normalized) bel(A|B) and $\oslash$ be $\oplus$ (Shafer 1976, Smets 1993). $\oslash$ doesn't have a counterpart in Shafer's presentation.

Let's consider two spaces $\Theta$ and X, we use bel$_X(.|\theta)$ to represent the belief function induced on the space X given $\theta \subseteq \Theta$. Suppose all we know about X is initially represented by the set $\{bel_X(.|\theta_i): \theta_i \in \Theta\}$. We only know the beliefs on X when we know which element of $\Theta$ holds. We do not know the belief on X when we only know that the prevailing element of $\Theta$ belongs to a given subset $\theta$ of $\Theta$. Under the requirement that the two pieces of evidence by which our belief function is induced are distinct and that the general likelihood principle is satisfied, Smets (1978, 1993) has derived the Disjunctive Rule of Combination (DRC) to build bel$_X(.|\theta)$ on X for any $\theta \subseteq \Theta$ and the Generalized Bayesian Theorem (GBT) to build bel$_\Theta(.|x)$ on $\Theta$ for any $x \subseteq X$.

**Theorem 1:** the Disjunctive Rule of Combination: $\forall \theta \subseteq \Theta, \forall x \subseteq X$,

$$m_X(x|\theta) = \sum_{\substack{\cup \ x_i = x \\ i: \theta_i \in \theta}} \prod_{i: \theta_i \in \theta} m_X(x_i|\theta_i) \tag{4.1}$$

$$pl_X(x|\theta) = 1 - \prod_{\theta_i \in \theta} (1 - pl_X(x|\theta_i)) \tag{4.2}$$

**Theorem 2:** the Generalized Bayesian Theorem: $\forall \theta \subseteq \Theta, \forall x \subseteq X$,

$$pl_\Theta(\theta|x) = 1 - \prod_{\theta_i \in \theta} (1 - pl_X(x|\theta_i)) \tag{5}$$

Now suppose there exists some a priori belief bel$_0$ over $\Theta$. By using Theorem 1 and 2, we can compute bel on X given bel$_0$ and $\{bel_X(.|\theta_i): \theta_i \in \Theta\}$:

**Theorem 3:** Suppose there exists some a priori belief bel$_0$ over $\Theta$ distinct from the belief induced by the set of conditional belief functions bel$_X(.|\theta_i): \theta_i \in \Theta$, then $\forall x \subseteq X$,

$$pl_X(x) = \sum_{\theta \subseteq \Theta} m_0(\theta) pl_X(x|\theta)$$
$$= \sum_{\theta \subseteq \Theta} m_0(\theta) \left(1 - \prod_{\theta_i \in \theta} (1 - pl_X(x|\theta_i))\right) \tag{6}$$

## 3. KNOWLEDGE REPRESENTATION USING BELIEF FUNCTIONS

Let U=$\{X_1, ..., X_n\}$ be a set of variables where each $X_i$ has its frame $\Theta_{X_i}$. Let A and B be two disjoint subsets of U, their frames are the product space of the frames of the variables they include. According to the notation of the previous section, a conditional belief function for B given A can be represented by bel$_{\Theta_B}(.|\theta)$ (bel$_B(.|\theta)$ for short) where $\theta \subseteq \Theta_A$, which means that we know the belief about B given the truth value of A is in $\theta$. In a valuation network, the same relationship between A and B is defined in a joint form on the space $\Theta_A \times \Theta_B$ ($\Theta_{A \cup B}$ or A×B for short). Look at the following example:

**Example 1:** Let A and B be two variables with frames $\Theta_A = \{a, \sim a\}$ and $\Theta_B = \{b, \sim b\}$ respectively. To represent a relation between A and B such as: if A=a then B=b with m=0.9, by a belief function in joint form, the rule is represented by a belief function on the space $\Theta = \{(a, b) (a, \sim b) (\sim a, b) (\sim a, \sim b)\}$, with masses: 0.9 on the subset $\{(a, b) (\sim a, b) (\sim a, \sim b)\}$, and 0.1 on $\Theta$, while by a belief

---

[1] We use "|" in place of "|" to enhance the non-normalization of our conditioning.

function in a conditional form, it is represented as: $m(\{b\}|a)=.9$, $m(\Theta_B|a)=.1$; $m(\Theta_B|\neg a)=1$. It can be represented by the following table:

Table 1: a Belief Function in a Conditional Form

|          | a   | ¬a  |
|----------|-----|-----|
| b        | 0.9 | 0   |
| $\Theta_B$ | 0.1 | 1   |

Obviously, the latter representation is more "natural" and "easy" for the user to provide and to understand. Generally, given two disjoint subsets $X, Y \subseteq U$, to represent a conditional belief function for Y given X, by a joint form, it needs $2^{|\Theta_X| \times |\Theta_Y|}$ elements in the worst case, while by a conditional form, it only needs $2^{|\Theta_X| + |\Theta_Y|}$ elements in the worst case.

Cano et al.(1993) has presented an axiomatic framework in directed acyclic networks which can propagate belief functions in the networks, and has given a definition for a non-informative belief function[2] in such framework represented by belief functions on the product space of two disjoint subsets. Shenoy (1993) has also shown the property of such belief functions in a valuation network. Let's first look at the concepts of projection, extension and marginalization:

**Definition 6:** Projection of configurations simply means dropping the extra coordinates. If X and Y are sets of variables, $Y \subseteq X$, and $x_i$ is an element of $\Theta_X$, then let $x_i^{\downarrow Y}$ denote the projection of $x_i$ to $\Theta_Y$. $x_i^{\downarrow Y}$ is an element of $\Theta_Y$. If x is a non-empty subset of $\Theta_X$, then the *projection of x to Y*, denoted by $x^{\downarrow Y}$, is obtained by $x^{\downarrow Y} = \{x_i^{\downarrow Y} \mid x_i \in x\}$. If y is a subset of $\Theta_Y$, then the *extension of y to X*, denoted by $y^{\uparrow X}$, is $y \times \Theta_{X \cdot Y}$ (It is also called *the cylinder set extension of y into X*).

**Definition 7:** Suppose m is a bba on X and suppose $Y \subseteq X \subseteq U$, $Y \neq \varnothing$. *The marginal of m for Y* denoted by $m^{\downarrow Y}$, is a bba on Y defined by $m^{\downarrow Y}(y) = \sum \{m(x) \mid x \subseteq \Theta_X, x^{\downarrow Y} = y\}$ for all subsets y of $\Theta_Y$.

**Definition 8:** Given two disjoint subsets $X, Y \subseteq U$ in the framework of Cano et al. (1993), let bel be a belief function defined on the space $\Theta_{X \cup Y}$. It is said that bel is a *non-informative belief function over X* iff $bel^{\downarrow X}$ is a vacuous belief function over X.

Intuitively, the belief function in definition 8 gives some information about variables in Y and their relationship with variables in X, but no information about X. This property is easy to verify when the belief is represented by a conditional form.

**Lemma 1:** $bel_Y(.|x)$: $x \subseteq \Theta_X$ is non-informative over X iff $bel_Y(.|x)$ is a normalized belief function for each

[2] Note that Shenoy (1993) and Cano et al. (1993) called this belief function "conditional belief function". We change the name to avoid confusion with the classical meaning of "conditional belief function".

$x \subseteq \Theta_X$. i.e., the representation $bel_Y(.|x)$: $x \subseteq \Theta_X$ is such a non-informative belief function over X.

Moreover, we can find that if a belief function bel defined on the space $\Theta_{X \cup Y}$ gives information only on the relationship of X and Y, but no information about either X or Y, then $bel^{\downarrow X}$ and $bel^{\downarrow Y}$ are both vacuous on X and Y respectively. That is to say, bel can be non-informative over either X or Y. The followings give the verification for the belief functions in the conditional form:

**Lemma 2:** Let $bel_Y(.|x)$, $x \subseteq \Theta_X$ be a conditional belief function for Y given X. It is non-informative over Y iff $bel_Y(.|\Theta_X)$ is a vacuous belief function over Y.

**Lemma 3:** If we only know the conditional belief function as $bel_Y(.|\theta_i)$, $\theta_i \in \Theta_X$, then it is non-informative over Y iff for each $y \subseteq \Theta_Y$, $\exists \theta_i \in \Theta_X$, such that $bel_Y(y|\theta_i)=0$.

In the following, we will show some relations between the belief functions represented in conditional form and in joint form. By using the rules of conditioning, every joint belief function can be transformed to a conditional form, but not every belief function in a conditional form can be transformed to a joint belief function. We say those that can not be transformed to joint beliefs are invalid. If it can be transformed, the joint form is not always unique. Smets (1993) has shown that when a conditional belief function is represented by $\{bel_Y(.|\theta_i): \theta_i \in \Theta_X\}$, we can always construct the joint belief from it.

**Lemma 4:** Let X and Y be two disjoint subsets of U. $m_{X \times Y}$ be a belief function on the product space $X \times Y$, representing a conditional belief function for Y given X. Then its conditional form $m_Y(.|x)$: $x \subseteq \Theta_X$ is obtained by:

$$m_Y(y|x) = \sum_{\substack{S \subseteq \Theta_Y \\ (S \cap x^{\uparrow (X \cup Y)})^{\downarrow Y} = y}} m_{X \times Y}(S) \qquad (7)$$

**Lemma 5:** If a belief function in a conditional form $bel_Y(.|x)$: $x \subseteq \Theta_X$ can be transferred to a joint belief, then it should satisfy $pl_Y(y|x_1) \leq pl_Y(y|x_2)$ if $x_1 \subseteq x_2 \subseteq \Theta_X$.
**Proof:** $pl_Y(y|x_1) = pl_{X \times Y}(y^{\uparrow X \cup Y}|x_1^{\uparrow X \cup Y}) =$
$pl_{X \times Y}(y^{\uparrow X \cup Y} \cap x_1^{\uparrow X \cup Y})$
$\leq pl_{X \times Y}(y^{\uparrow X \cup Y} \cap x_2^{\uparrow X \cup Y}) = pl_{X \times Y}(y^{\uparrow X \cup Y}|x_2^{\uparrow X \cup Y})$
$= pl_Y(y|x_2)$. \hfill QED

**Example 2:** Let A and B be two variables with frames $\Theta_A = \{a, \neg a\}$ and $\Theta_B = \{b, \neg b\}$ respectively. Let $\Theta = \{ab, a \neg b, \neg ab, \neg a \neg b\}$ be shortly denoted by $\{1, 2, 3, 4\}$, then subset $\{ab, a \neg b\}$ can be denoted by 12, for example. Consider a belief function $bel_1$ on $\Theta$: $m(14)=m(23)=0.1$, $m(123)=m(124)=m(134)=m(234)=0.1$ and $m(1234)=0.4$, by applying lemma 4, we have its corresponding conditional belief function for B given A shown in table2.a; However, for another belief function on $\Theta$: $m(23)=0.2$, $m(134)=m(124)=0.2$ and $m(1234)=0.4$, its corresponding conditional form by applying lemma 4 is shown in Table 2.b. Comparing the two tables, we can find, therefore, that two different joint belief function might be transferred to the same conditional form.

Table 2.a: Belief Function in Conditional Form for $bel_1$

|  | a | ~a | $\Theta_A$ |
|---|---|---|---|
| b | $m(14)+m(134)$ $=.1+.1=.2$ | $m(23)+m(123)$ $=.1+.1=.2$ | 0 |
| ~b | $m(23)+m(234)$ $=.1+.1=.2$ | $m(14)+m(124)$ $=.1+.1=.2$ | 0 |
| $\Theta_B$ | $m(123)+m(124)$ $+m(1234) =$ $.1+.1+.4=.6$ | $m(134)+m(234)$ $+m(1234) =$ $.1+.1+.4=.6$ | $m(14)+m(23)+m(123)$ $+m(124)+m(134)+$ $m(234)+m(1234)=1$ |

Table 2.b: Belief Function in Conditional Form for $bel_2$

|  | a | ~a | $\Theta_A$ |
|---|---|---|---|
| b | $m(134) = .2$ | $m(23) =.2$ | 0 |
| ~b | $m(23) = .2$ | $m(124) =.2$ | 0 |
| $\Theta_B$ | $m(124)+m(1234)$ $=.2+.4 =.6$ | $m(134)+m(1234)$ $= .2+.4 =.6$ | $m(23)+m(124)+$ $m(134)+m(1234)=1$ |

**Lemma 6:** Suppose X and Y are two disjoint subsets of U. For each $x_i \in \Theta_X$, let $bel_Y(.|x_i)$ denote a belief function on $\Theta_Y$. Given these belief functions, we can construct the belief function on $\Theta_{X \cup Y}$ as follows (Smets 1993):

Let $bel_{X \cup Y}$ be the resulting belief function on $\Theta_{X \cup Y}$, called the *ballooning extension of $bel_Y(./x_i)$.* Let $a \subseteq \Theta_{X \cup Y}$ and $y_i$ be the projection of $a \cap \{x_i\}^{\uparrow (X \cup Y)}$ for Y. Then

$$m_{X \cup Y}(a) = \prod \{m_Y(y_i|x_i) | x_i \in \Theta_X\} \qquad (8)$$

## 4. REASONING WITH CONDITIONAL BELIEFS

In this paper, we use the network proposed by Smets (1993) for the propagation of beliefs. Graphically, the network is a directed acyclic graph (dag) as defined in Pearl (1988) for the Bayesian networks, shown in Figure 1. A graph $G = (M, E)$, where M are the finite sets of nodes and E are the sets of edges, is said to be a dag when there is no path $n_1 n_2 ... n_k$ such that $(n_i, n_{i+1}) \in E$ $(1 < i \le k-1)$ and $n_1 = n_k$. However, the conditional beliefs are defined in a different way. In our network, each edge represents a conditional relation between the two nodes it connects. In order to distinguish these two kinds of networks, we call ours ENC, which means an evidential network with conditional belief functions. We also assume that, for each conditional belief function for Y given X, all we know about Y given X is initially represented by the set $\{bel_Y(.|x_i): x_i \in \Theta_X\}$. For example, in Figure 1, edge (A, B) represents a conditional belief function for the node B given A, represented by $bel_B(.|a_i): a_i \in \Theta_A$.

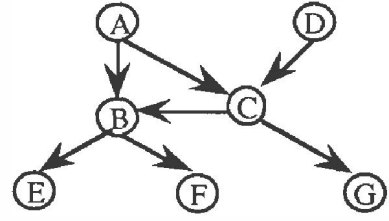

Figure 1: an Evidential Network with Conditional Belief Functions

One main object of reasoning process in evidential network is to compute the marginal distributions for some variables. We use $BEL_X$ to denote the marginal for variable X, $bel_{0X}$ the a priori belief for X. Due to the DRC and the GBT, given two variables X and Y, and the conditional belief $bel_Y(.|x_i)$: $x_i \in \Theta_X$, we could compute and store $bel_Y(.|x): x \subseteq \Theta_X$ and $bel_X(.|y): y \subseteq \Theta_Y$ in the preprocess, which might be useful for speeding up the computation in the propagation. Now, we are ready to give the inference algorithm: Given an ENC represented by $G=(M, E)$.

Case1: propagating beliefs in polytrees, i.e., there is only one (undirected) path between any of two nodes in the network:

Propagation algorithm can be regarded as a message-passing scheme: for each node X in the network, its marginal $BEL_X$ is computed by combining all the messages from its neighbors $N_X=\{Y(\in M)|(X, Y)\in E$ or $(Y, X)\in E\}$ and its own a priori belief $bel_{0X}$. i.e.,

$$BEL_X= bel_{0X} \otimes (\otimes \{M_{Y \to X} \mid Y \in N_X\}) \qquad (9)$$

where the message $M_{Y \to X}$ is a belief function on X, and it can be represented by $bel_{Y \to X}$ or $m_{Y \to X}$, and is computed by: for any $x \subseteq \Theta_X$,

$$bel_{Y \to X}(x) = \sum_{y \subseteq \Theta_Y} m_X(x|y) \cdot bel_{N_Y/X \to Y}(y) \text{ where}$$

$$bel_{N_Y/X \to Y}=bel_{0Y} \otimes (\otimes \{bel_{Z \to Y}|Z \in N_Y \ \& \ Z \neq X\}) \qquad (10)$$

Case 2: If there exist any undirected loops in the network, then some nodes needed to be merged to make the network acyclic, resulting in a new polytree $G'=(M', E')$, where some nodes in $G'$ might be a subset of the nodes in G, we call this kind of node a *merged node*. For any merged node v in $G'$, there might be a belief function $R_V$ obtained by the ballooning extension of conditional beliefs. Figure 2 illustrates two examples for this process:

In Figure 2.a, the loop is absorbed by merging nodes B and C, the resulting graph is shown in 2.b where $D=\{B, C\}$, and new conditional belief function $bel_D(.|a_i)$ is obtained by combining $bel_B(.|a_i)$ and $bel_C(.|a_i)$ on the space $\Theta_D=\Theta_{B \cup C}$: $\forall a_i \in \Theta_A, d \subseteq \Theta_D$,

$$m_D(d|a_i)= \sum_{b^{\uparrow\{B,C\}} \cap c^{\uparrow\{B,C\}}=d} m_B(b|a_i) \cdot m_C(c|a_i) \qquad (11)$$

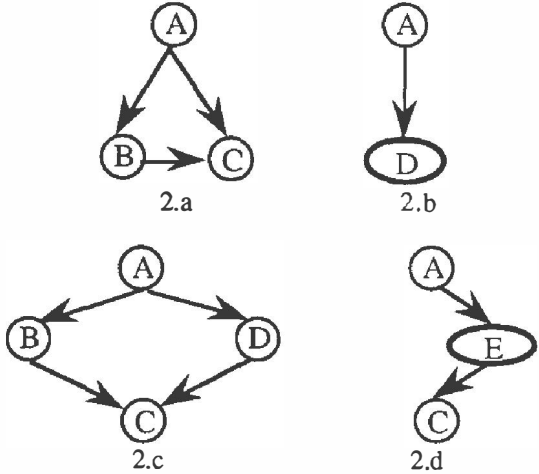

Figure 2: Examples of Absorbing Loops in the ENCs

Obviously, $bel_D(d|a_i)$ is normalized iff $bel_B(.|a_i)$ and $bel_C(.|a_i)$ are normalized since the subset $b^{\uparrow\{B,C\}} \cap c^{\uparrow\{B,C\}}$ can never be an empty set. Moreover, the conditional belief function between B and C becomes $R_D$ in Figure 2.b obtained by the ballooning extension of $bel_B(.|c_i)$ applying eq. (8). Thus $R_D$ is a belief function on $\Theta_D$.

Figure 2.c is another example of ENC with a loop. In this case, we merge B and D, resulting in the graph shown in 2.d, where $E = \{B, D\}$. $bel_E(.|a_i)$ is obtained by combining $bel_B(.|a_i)$ and $bel_D(.|a_i)$ on $\Theta_E = \Theta_{B \cup D}$ using eq. (11). As for $bel_C(.|e)$: $e \subseteq \Theta_E$, we compute it for three cases:

1) for any $e_t = (b_i, d_j) \in \Theta_E$,
$$m_C(c|e_t) = \sum_{s_1 \cap s_2 = c} m_C(s_1|b_i) \cdot m_C(s_2|d_j); \quad (12.1)$$

2) for $e \subseteq \Theta_E$, if e can be represented by $b \times d$, where $b \subseteq \Theta_B$, $d \subseteq \Theta_D$,
$$m_C(c|e) = \sum_{s_1 \cap s_2 = c} m_C(s_1|b) \cdot m_C(s_2|d) \quad (12.2)$$

where $m_C(.|b)$ and $m_C(.|d)$ are obtained from $m_C(.|b_i)$ and $m_C(.|d_j)$ respectively by applying the DRC as shown in equations (4);

3) for any other $e \subseteq \Theta_E$, we first construct a conditional belief function $bel_{C \cup D}(.|b_i)$ from $m_C(.|b_i)$ such that

$m_{C \cup D}(s|b_i) = m_C(c|b_i)$

where $s = c^{\uparrow\{C,D\}} \cap ((e \cap b_i^{\uparrow E})^{\downarrow\{D\}})^{\uparrow\{C,D\}}$, let $bel_{C \cup D}^b$ be the belief function resulting from the ballooning extension of $m_C(.|d_i)$, then

$$m_C(c|e) = (bel_{C \cup D}^b \oslash (\oslash \{bel_{C \cup D}(.|b_i)|b_i \in \Theta_B\}))^{\downarrow\{C\}} \quad (12.3)$$

Alternatively, $bel_C(.|e)$: $e \subseteq \Theta_E$ can be computed by first combining the ballooning extensions of the two conditional beliefs $bel_C(.|b_i)$ and $bel_C(.|d_j)$ on the space $\Theta_{B \cup C}$ and $\Theta_{C \cup D}$, then using equation (7) to transfer the resulting belief in a conditional form $bel_C(.|e):e \subseteq \Theta_E$ and

$bel_E(.|c):c \subseteq \Theta_C$. However, this takes more space for the computation.

Since there is no direct relation between B and D, $R_E$ is a vacuous belief function.

After transforming the network to an acyclic one, we then use a similar algorithm in case1 for the propagation: Suppose each node X in G' is a subset and has a $R_X$. Thus, for any non-merged node, it is a singleton, and $R_X$ is a vacuous belief function. Then the computation is as following: for any node $A = \{X_1, ..., X_t\}$ in G',

$$Bel_A = R_A \oslash (\oslash \{M_{Y \to A} | Y \in N_A\}) \text{ and} \quad (13.1)$$
$$BEL_{X_i} = bel_{0X_i} \oslash (Bel_A \oslash$$
$$(\oslash \{bel_{0X_j} | X_j \in A, X_j \neq X_i\}))^{\downarrow X_i} \quad (13.2)$$

the message $M_{Y \to A}$ from Y to A is computed by: for $Y = \{Y_1, ...Y_n\}$,

$$bel_{Y \to A}(a) = \sum_{y \in \Theta_Y} m_A(a|y) \cdot bel_{N_Y/A \to Y}(y) \text{ where}$$

$$bel_{N_Y/A \to Y} = R_Y \oslash (\oslash \{bel_{0Y_j} | Y_j \in Y\}) \oslash \{M_{Z \to Y} | Z \in N_Y, Z \neq A\})) \quad (13.3)$$

Although the above representation and propagation algorithm are for the networks which only have binary relations between the nodes, it could be generalized to the case where relations are for any number of nodes. In the rest of this section, we will show some special cases where using ENC can reduce the computation.

**Definition 9:** Let X, Y be two nodes in ENC, where $\Theta_X = \{x_1, ...x_p\}$, $\Theta_Y = \{y_1, ...y_q\}$. Suppose there is an edge (X, Y) representing a conditional belief for Y given X: $bel_Y(.|x_i)$: $x_i \in \Theta_X$ such that $m(\Theta_Y|x_i) < 1$ for $i = 1, ..., t(<p)$ and $m(\Theta_Y|x_j) = 1$ for $j = t+1, ...p$. We say the elements $x_i$'s ($i \leq t$) are *relevant* to Y and $x_j$'s ($t < j \leq p$)*irrelevant* to Y.

This kind of relationship exists commonly in the diagnosis problems and rule-based systems. In Example 1, we say that a is relevant to B, but ~a irrelevant to B. Intuitively, it means that given some knowledge on a, we can induce knowledge about B, but no matter what we know about ~a, we can't induce any knowledge about B. Thus we say ~a is irrelevant to B.

**Lemma 7:** Given two variables X, Y and the conditional bel on Y given X, suppose $\Phi = \{x_{t+1}, .., x_p\}$ is irrelevant to Y. Then for any subset S of $\Theta_X$, if $S \cap \Phi \neq \emptyset$, then $m_Y(\Theta_Y|S) = 1$.

**Proof:** The result can be derived directly by applying the GBT.                QED

**Lemma 8:** Given two variables X, Y and a conditional bel on Y given X, suppose $\Phi = \{x_{t+1}, .., x_p\}$ is irrelevant to Y. Assume we have some belief $bel_0$ on Y, by theorem 1-3, we can compute the belief of X. If $m_X(S) \neq 0$, then $S \supseteq \Phi$.

**Proof:** From lemma 7 we have, $\forall x \subseteq \Theta_X$, if $x \cap \Phi \neq \emptyset$, $m_Y(\Theta_Y|x) = 1$, i.e., $pl_Y(y|x) = 1$ for $\forall y \subseteq \Theta_Y$. Then, by equations (5) and (6), $pl_X(x|y)=pl_Y(y|x)=1$ for such x. Thus by equation (7), we have

$$pl_X(x) = \sum_{y \subseteq \Theta_Y} m_0(y) \cdot pl_X(x|y) = \sum_{y \subseteq \Theta_Y} m_0(y) = 1.$$

Therefore, $\forall S \subseteq \Theta_X$, if $m_X(S) \neq 0$, S should contain any element of $\Phi$, i.e. $S \supseteq \Phi$.    QED

From lemma 7 & 8, we can simplify the computation for some special cases of ENC, shown in Figure 3, where in 3.a, $G_i$ is a group (set) of variables and suppose some elements $\Phi_i$ of A are irrelevant to each variable $X_i$. Figure 3.b shows detail in each $G_i$. To describe the computation, let's begin by recalling the concept of partition:

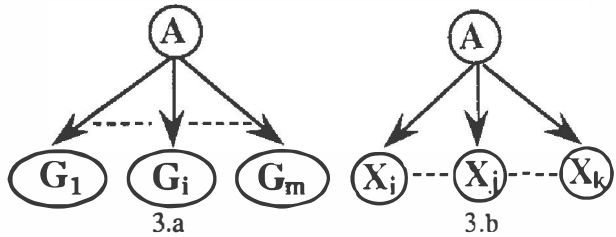

3.a                    3.b

Figure 3: an Example of Some Special Case of ENC

**Definition 10:** Let $\Theta = \{\theta_1, ..., \theta_p\}$ be a frame of discernment. A set $\mathcal{P}_\Theta$ of subsets of $\Theta$ is *a partition* of $\Theta$ if the elements in $\mathcal{P}_\Theta$ are all non-empty and disjoint and their union is $\Theta$. We also call $\mathcal{P}_\Theta$ *a coarsening* of $\Theta$ and $\Theta$ *a refinement* of $\mathcal{P}_\Theta$.

From the definition, we have $\forall \theta_i \in \Theta$, $\exists x_j \in \mathcal{P}_\Theta$ which is a mapping of $\theta_i$. We denote such mapping by $\Lambda(\theta_i)=x_j$. $\forall \theta \subseteq \Theta$, $\Lambda(\theta)=\{\Lambda(\theta_i)|\theta_i \in \Theta\}$. Let $bel_1$ be a belief function on $\Theta$, then the belief $bel_2$ on $\mathcal{P}_\Theta$ induced by $bel_1$, say, by coarsening, is obtained by: $\forall x \subseteq \mathcal{P}_\Theta$,

$$m_2(x) = \sum_{\Lambda(\theta)=x} m(\theta) \qquad (14.1)$$

Let $bel_2$ be a belief function on $\mathcal{P}_\Theta$, $bel_1$ on $\Theta$ induced by $bel_2$, say, by refinement, is obtained by: $\theta \subseteq \Theta$, $x \subseteq \mathcal{P}_\Theta$,

$$m(\theta) = m(x) \qquad (14.2)$$

where $\theta = \cup \{\theta' \mid \Lambda(\theta') = x\}$

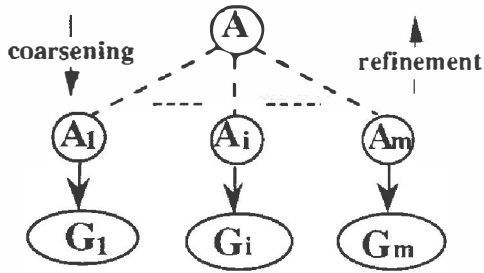

Figure 4: Two-Level Structure for the Computation in the Network shown in Figure 3

Given a network as shown in Figure 3, we can represent it as a two-level structure shown in Figure 4. Each $A_i$ has a frame $\Theta_{A_i}$ which is a partition of $\Theta_A$ such that $\forall a_k \in \Theta_A$, $\Lambda(a_k)=s_i$ if $a_k \in (\cap \{\Phi_j \mid X_j \subseteq G_i\})$, otherwise $\Lambda(a_k)=\{a_k\}=a_k$. In fact, $s_i=\cap\{\Phi_j \mid X_j \subseteq G_i\}$. Each $\boxed{A_i \rightarrow G_i}$ part can be regarded as a local sub-network, and the belief functions passed between A and $A_i$ are performed by refinement and coarsening between the two frames. Let's look at the following example:

**Example 3:** Suppose we have 4 variables in the network (shown in Figure 5.a): A, X, Y and Z. Their frames are: $\Theta_A = \{a_1, a_2, a_3, a_4, a_5\}$, $\Theta_X = \Theta_Y = \Theta_Z = \{+, -\}$. The relations among them are represented by conditional belief functions in Table 3.

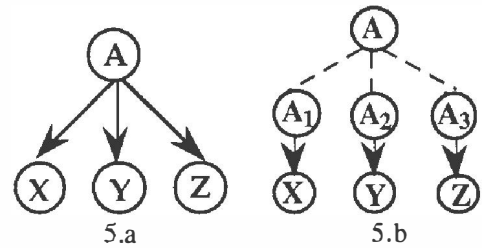

5.a                    5.b

Figure 5: An Example of ENC

Table 3: Conditional belief functions for example 3

$m_X(x|a_i)$ i=1,.., 5      $m_Y(y|a_i)$ i=1,.., 5

| | a1 | a2 | a3 | a4 | a5 | | a1 | a2 | a3 | a4 | a5 |
|---|---|---|---|---|---|---|---|---|---|---|---|
| + | .9 | .7 | 0 | 0 | 0 | + | 0 | .7 | .2 | .4 | 0 |
| - | 0 | .3 | 0 | 0 | 0 | - | 0 | .3 | .6 | .1 | 0 |
| $\Theta$ | .1 | 0 | 1 | 1 | 1 | $\Theta$ | 1 | 0 | .2 | .5 | 1 |

$m_Z(z|a_i)$ i=1,.., 5

| | a1 | a2 | a3 | a4 | a5 |
|---|---|---|---|---|---|
| + | 0 | 0 | 0 | .6 | .9 |
| - | 0 | 0 | 0 | .3 | 0 |
| $\Theta$ | 1 | 1 | 1 | .1 | .1 |

Now suppose that we have some observation about X and Z: $m_{0X}(\{+\})=.8$, $m_{0X}(\Theta_X)=.2$; $m_{0Z}(\{-\})=1$. To compute the marginal for A, if we use the joint belief for the relation A and X, then the combination is performed on the product space $\Theta_A \cup_X$ and $\Theta_A \cup_Z$; if we use the conditional belief represented in the above tables, the computation is performed on the frame $\Theta_A$, which is more efficient. Moreover, if we use the result of lemma 8, the computation can be simplified further. The following steps illustrate such computation:

1. transform the network in 5.a to the network shown in Figure 5.b where each $\Theta_{A_i}$ is a partition of $\Theta_A$:

$\Theta_{A_1}=\{a_1, a_2, s_1\}$ $(s_1=\{a_3,a_4,a_5\})$ $\Theta_{A_2}=\{a_2, a_3, a_4, s_2\}$, and $\Theta_{A_3}=\{a_4, a_5, s_3\}$. $bel_X(.|a_j)$: $a_j \in \Theta_{A_1}$ is obtained from $bel_X(.|a_j)$: $a_j \in \Theta_A$, $bel_X(.|s_1)$: $s_1 \in \Theta_{A_1}$ is obtained by applying the DRC, Symmetrically, we can get the other two conditional beliefs. The resulting conditional beliefs are shown in Table 4:

2. Using the DRC to compute $bel_{A_1}(.|x)$ and $bel_{A_3}(.|z)$.

3. Using Theorem 2-3 to compute $bel_{A_i}$: i=1,2,3. $bel_{A_2}$ is vacuous by lemma 3;
   $m_{A_1}(\{a_1, s_1\})=.24$, $m_{A_1}(\Theta_{A_1})=.76$; and
   $m_{A_3}(\{s_3\})=.54$, $m_{A_3}(\{a_4,s_3\})=.36$, $m_{A_3}(\{a_5,s_3\})=.06$, $m_{A_3}(\Theta_{A_3})=.04$.

4. compute the above two beliefs on the frame $\Theta_A$ by refinement and combine them, we get our desired result.

Table 4: Conditional Beliefs Induced from Table 3 for the Partition of $\Theta_A$

| $m_X(x|a_i)$: $a_i \in \Theta_{A_1}$ | | | | $m_Y(y|a_i)$: $a_i \in \Theta_{A_2}$ | | | | $m_Z(z|a_i)$: $a_i \in \Theta_{A_3}$ | | |
|---|---|---|---|---|---|---|---|---|---|---|
| | $a_1$ | $a_2$ | $s_1$ | | $a_2$ | $a_3$ | $a_4$ | $s_2$ | $a_4$ | $a_5$ | $s_3$ |
| + | .9 | .7 | 0 | + | .7 | .2 | .4 | 0 | + | .6 | .9 | 0 |
| - | 0 | .3 | 0 | - | .3 | .6 | .1 | 0 | - | .3 | 0 | 0 |
| $\Theta$ | .1 | 0 | 1 | $\Theta$ | 0 | .2 | .5 | 1 | $\Theta$ | .1 | .1 | 1 |

Obviously, this computation is more efficient since in step 2 and 3, the computation is taken on the frame $\Theta_{A_i}$ which is smaller than $\Theta_A$.

Moreover, if the network has the properties defined as below, we can also simplify the computation for each sub-network shown in Figure 3.b.

**Definition 11:** Let X, Y and A be three nodes in an ENC, where $\Theta_X=\{x_1,...,x_p\}$, $\Theta_Y=\{y_1,...,y_q\}$ and $\Theta_A=\{a_1,...,a_t\}$. Suppose we have $bel_X(.|a_i)$ and $bel_Y(.|a_i)$ for $a_i \in \Theta_A$. Let $\Phi_X$, $\Phi_Y \subseteq \Theta_A$ be the sets of irrelevant elements for X and Y respectively. $\Phi_X \neq \emptyset$, $\Phi_Y \neq \emptyset$. X and Y are *unrelated* through A, denoted by $u(X,Y,A)$, if $\Phi_X$ and $\Phi_Y$ satisfy:

(1) $\Phi_X \cap \Phi_Y \neq \emptyset$ ; or

(2) $\Phi_X \cap \Phi_Y = \emptyset$, $\Phi_X \cup \Phi_Y = \Theta_A$, $bel_X(.|\bar{\Phi}_X)$ or $bel_Y(.|\bar{\Phi}_Y)$ obtained from $bel_X(.|a_i)$ or $bel_Y(.|a_i)$ by the DRC is vacuous.

This relation can also be extended to the two disjoint subsets, where $\Phi_A$: $A \subseteq U$ is defined as: $\Phi_A = \cap \{\Phi_X |$ $X \in A\}$.

**Lemma 9:** Let X, Y and A be defined as in definition 11. Suppose we have $bel_X(.|a_i)$ and $bel_Y(.|a_i)$ for $a_i \in \Theta_A$, but no a priori belief on A. Now suppose we have observations about X, then $BEL_Y$ is vacuous if X and Y are unrelated.

**Proof:** This can be proved by applying the results of lemma 7 and 8.    QED

Now let's consider the computation for the network shown in Figure 6.a. Let X, Y, A, $\Phi_X$ and $\Phi_Y$ be defined as in definition 11. Assume X and Y are binary variables. Suppose we have conditional beliefs for X and for Y given each element of $\Theta_A$, $u(X,Y,A)$ and $\bar{\Phi}_X \cap \bar{\Phi}_Y = \emptyset$, the conditional belief for Y given X $bel_Y(.|x_i)$: i=1,2 is such that $bel_Y(.|\Theta_A)$ obtained from $bel_Y(.|x_i)$ is vacuous. If all the a prori beliefs for the variables are vacuous, the propagation result would be vacuous. Now suppose we have observations about X: $X=x_i$, then the network in 6.a is equivalent to the one shown in 6.b (Proofs can be found in (Xu and Smets 1994)).

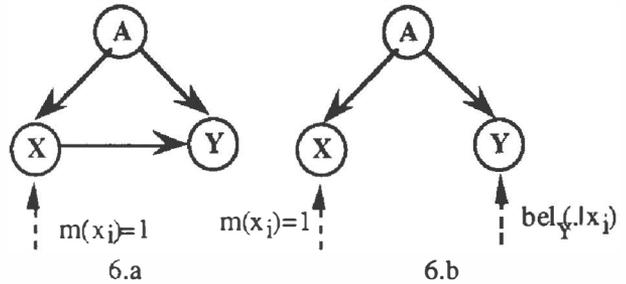

Figure 6: an ENC where X and Y Are Unrelated through A

The computation of Figure 6.b is obvious by applying the DRC and the GBT. Let $m_{A|x_i}$ denote the resulting belief for A. Furthermore, if the observation about X is any kind of belief function, we can compute $BEL_A$ for A as follows: $\forall a \subset \Theta_A$,

$$m(a) = \sum_{x_i \in \Theta_X} m_{0x}(x_i)m_{A|x_i} \quad \text{and}$$

$$m(\Theta_A) = 1 - \sum_{a \subset \Theta_A} m(a).$$

If there exists some a priori belief for A: $bel_{0A}$, then the marginal for A is computed by the combination of the above resulting belief function and $bel_{0A}$.

This method avoids the computation on the frame of $\Theta_{A \cup X \cup Y}$ for the case where the conditional belief is represented in joint form and avoids merging nodes to compute $bel(a|x,y)$ and $bel(x,y)$ as described in the beginning of the section, thus it simplifies the computation for such kind of network. This method can also be extended for the case shown in Figure 7.a, under the condition that $u(G_{-1}, G_{+1} \cup \{X_j\}, A)$, $u(G_{+1}, G_{-1} \cup \{X_j\}, A)$ and the intersection of the relevant elements of each pair of $G_{-1}$, $G_{+1}$ and $X_j$ is empty. The network in 7.a is equivalent to the one in 7.b (Proofs can be found in (Xu and Smets 1994)). If $X_i$'s inside G have similar relationship, we can iteratively use the scheme described above to compute the marginal for A.

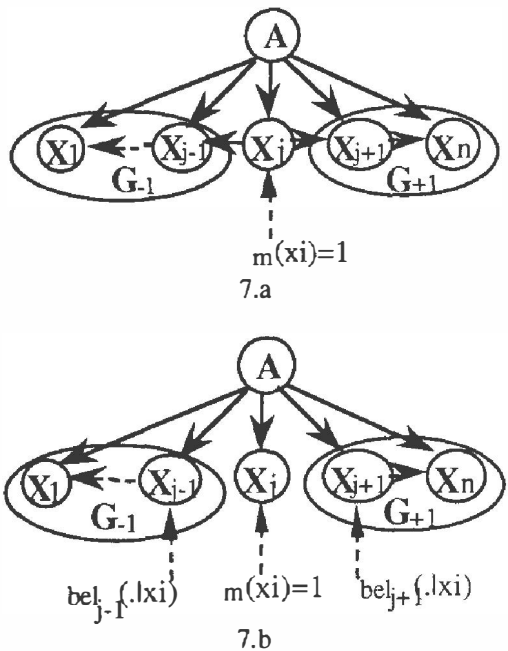

m(xi)=1

7.a

bel$_{j\text{-}f}$(.lxi)     m(xi)=1     bel$_{j+f}$(.lxi)

7.b

Figure 7: A General Case of ENC Whose Computation
Could be Simplified

## 5. CONCLUSIONS

We have presented an evidential network (ENC) which
uses conditional belief functions for the knowledge
representation and reasoning. In the paper, we have
compared some relations between the representation by
joint belief and by conditional belief and we have found
that the conditional form is more natural and it takes less
space. We also provide an algorithm for reasoning in
ENCs. The presented algorithm of reasoning is only for
the network where all the relations are binary, the
extension of the algorithm to a general case will be
studied in the future work. Although we have compared
the computational complexity of ENC and the network
using joint beliefs in a general case, we have shown that
in some special cases, the computation of ENC can be
simplified and is more efficient than the network using
joint beliefs. The advantage of simplified computation in
such networks can be shown in the example abstracted
from (Xu et al. 1993).

### Acknowledgments

This research work has been partially supported by the
Action de Recherches Concertées BELON funded by a
grant from the Communauté Française de Belgique and
the ESPRIT III, Basic research Project Action 6156
(DRUMS II) funded by a grant from the Commission of
the European Communities. The authors are grateful to
anonymous reviewers for their comments and suggestions.
The first author would like to acknowledge the support by
a grant of IRIDIA, Université libre de Bruxelles.

### References

Cano J., Delgado M., and Moral S. "An Axiomatic
Framework for Propagating Uncertainty in Directed
Acyclic Networks" *IJAR*, 8:253-280, 1993.

Dubios D. and Prade H. "A set theoretical view of belief
functions" Int. J. Gen. Systems, 12:193-226, 1986.

Pearl J., *Probabilistic Reasoning in Intelligent Systems:
Networks of Plausible Inference*, Los Altos, CA,
Morgan Kaufmann, 1988.

Shafer G., *A Mathematical Theory of Evidence* Princeton
University Press, 1976.

Shafer G. and Shenoy P. P. "Local Computation in
Hypertrees" Working Paper No. 201, School of
Business, University of Kansas, Lawrence, KS, 1988.

Shenoy P. P. and Shafer G. "Axioms for probability and
belief functions propagation", *Uncertainty in Artificial
Intelligence 4*, (Shachter R. D., Levitt T. S., Kanal L.
N. and Lemmer J. F. eds.), North-Holland, Amsterdam,
159-198, 1990.

Shenoy P. P., "Valuation-Based Systems: A framework
for managing uncertainty in expert systems" *Fussy
logic for the Management of Uncertainty* (L. A. Zadeh
and J. Kacprzyk eds.), John Wiley & Sons, NewYork.
pp. 83-104, 1992.

Shenoy P. P., "Valuation Networks and Conditional
Independence" *Proc. 9th Uncertainty in AI* (M. P.
Wellman et. al. eds.), San Mateo, Calif.: Morgan
Kaufmann, pp.191-199, 1993.

Smets Ph. "Un modèle mathématico-statistique simulant
le processus du diagnostic médical", Doctoral
dissertation, Université libre de Bruxelles, 1978.Smets
Ph., "Belief Functions" *Non Standard Logics for
Automated Reasoning* (Smets Ph., Mamdani A.,
Dubois D. and Prade H. eds.), Academic Press, London,
253-286, 1988.

Smets Ph. "The combination of evidence of the
transferable belief model" *IEEE-Pattern Analysis and
Machine Intelligence*, 12:447-458, 1990.

Smets Ph., "Belief Functions: the Disjunctive Rule of
Combination and the Generalized Bayesian Theorem"
*IJAR*, Vol. 9, No. 1 pp. 1-35, 1993.

Xu H., Hsia Y. and Smets Ph., "A belief function based
decision support system" *Proc. 9th Uncertainty in AI*
(M. P. Wellman et. al. eds.), San Mateo, Calif.:
Morgan Kaufmann, pp.535-542, 1993.

Xu H. and Smets Ph., "Evidential reasoning with
conditional belief functions" Technical Report
TR/IRIDIA/94-5, Université libre de Bruxelles,
Belgium, 1994.